\documentclass{article}

\usepackage{multirow}
\usepackage{arxiv}
\usepackage{float} 
\usepackage[utf8]{inputenc} 
\usepackage[T1]{fontenc}    
\usepackage{hyperref}       
\usepackage{url}            
\usepackage{booktabs}       
\usepackage{amsfonts}       
\usepackage{nicefrac}       
\usepackage{microtype}      
\usepackage{lipsum}		
\usepackage{graphicx}
\usepackage{natbib}
\usepackage{doi}
\usepackage{tabularx} 
\usepackage{booktabs} 
\usepackage{algorithm}
\usepackage{algpseudocode}
\usepackage{amsmath}

\title{Advancing Meteorological Forecasting: AI-based Approach to Synoptic Weather Map Analysis}

\author{Yo-Hwan Choi \and Seon-Yu Kang\and Minjong Cheon\thanks{Corresponding author. Email: \texttt{jmj2316@kaist.ac.kr}}}

\date{
    \textsuperscript{1}Korea Power Exchange 625 Bitgaram-ro, Naju-si, Jeollanam-do, Republic of Korea.  \\
    \textsuperscript{2}Seoul National University, 1 Gwanak-ro, Gwanak-gu, Seoul, Republic of Korea. \\
    \textsuperscript{3}Korea Advanced Institute of Science and Technology, 91, Daehak-ro, Yuseong-gu, Daejeon, Republic of Korea	
}

\renewcommand{\shorttitle}{\textit{arXiv} Template}

\hypersetup{
pdftitle={A template for the arxiv style},
pdfsubject={q-bio.NC, q-bio.QM},
pdfauthor={Minjong},
pdfkeywords={First keyword, Second keyword, More},
}

\begin{document}
\maketitle

\begin{abstract}
As global warming increases the complexity of weather patterns; the precision of weather forecasting becomes increasingly important. Our study proposes a novel preprocessing method and convolutional autoencoder model developed to improve the interpretation of synoptic weather maps. These are critical for meteorologists seeking a thorough understanding of weather conditions. This model could recognize historical synoptic weather maps that nearly match current atmospheric conditions, marking a significant step forward in modern technology in meteorological forecasting. This comprises unsupervised learning models like VQ-VQE, as well as supervised learning models like VGG16, VGG19, Xception, InceptionV3, and ResNet50 trained on the ImageNet dataset, as well as research into newer models like EfficientNet and ConvNeXt. Our findings proved that, while these models perform well in various settings, their ability to identify comparable synoptic weather maps has certain limits. Our research, motivated by the primary goal of significantly increasing meteorologists' efficiency in labor-intensive tasks, discovered that cosine similarity is the most effective metric, as determined by a combination of quantitative and qualitative assessments to accurately identify relevant historical weather patterns. This study broadens our understanding by shifting the emphasis from numerical precision to practical application, ensuring that our model is effective in theory practical, and accessible in the complex and dynamic field of meteorology.\end{abstract}


\section{Introduction}
Accurate weather prediction has become increasingly important in an era of rising global warming \cite{1}. Weather predictions are directly tied to human existence including business, industry, transportation, and cultural-leisure activities globally \cite{2}. Weather forecasting is becoming increasingly difficult but still necessary due to the severity of climate change \cite{3}. Extreme weather events, formerly thought to be uncommon, are becoming more frequent and intense due to substantial changes in our atmosphere \cite{4}. More advanced and precise weather prediction methods must be developed to protect companies and communities from the unpredictable effects of climate change. FourCastNet, ClimaX, GraphCast, Fengwu, Fuxi, Pangu, and KARIN are representative models that are transforming weather prediction \cite{5}\cite{6}\cite{7}\cite{8}\cite{9}\cite{10}. To improve forecast accuracy, these complicated models include input variables from ERA5, a massive database of meteorological observations \cite{11}. 

However, despite prompt advances in AI-based weather forecasting models, there is a considerable vacuum in models created particularly for synoptic weather map analysis \cite{12}. Synoptic weather maps are important in meteorology because they provide a thorough picture of weather conditions across a large horizontal distance of around 1,000 km \cite{13}. These maps offer a comprehensive weather overview across a sizable region at a certain time. For weather forecasts to be accurate and dependable, synoptic maps must be thoroughly analyzed since they condense the complexity of atmospheric dynamics into an approachable format \cite{14}. Meteorologists at the beginner stage are trained by comparing current weather data against historical maps. This process is based on analyzing a vast archive of past weather patterns. However, traditional manual work is labor-intensive and prone to errors, and Galen et al. also proposed the importance of adapting new methodologies for meteorologists \cite{15}. Our paper proposes a Convolutional Autoencoder-based model to overcome these challenges and enhance forecasting practices. This model efficiently identifies historical weather maps like current conditions, representing a significant technological advancement in meteorology. It promises to revolutionize how meteorologists access and analyze historical data, potentially greatly improving the accuracy and reliability of weather forecasts by integrating traditional meteorological expertise with modern technological efficiency. 

Our research addresses several key objectives by focusing on both the application and effectiveness of our proposed model, at first, we aim to determine the most appropriate metric for assessing the performance of our model in identifying historical weather patterns that match current conditions.  Secondly, we propose a data preprocessing technique that enhances the model's capability to process meteorological data. This technique is designed to optimize the quality of input data, which could help Autoencoder identify patterns accurately. Thirdly, we perform a comparative analysis of our model against a range of pretrained models based on various metrics. Lastly, we also conduct a qualitative approach based on interpreting the model’s outputs, which could help cross-check the validity of the result.

\section{Materials and Methods}
Ahn et al. offered a deep learning strategy that uses convolutional autoencoders (CAEs) and convolutional neural networks (CNNs) to help weather forecasters detect previous weather maps like current ones. By converting satellite photos into low-dimensional vectors, the model detects dates with weather circumstances similar to a particular reference date using Euclidean distances between their latent feature vectors. For the given work, the CAE model outperformed typical pre-trained CNN models in terms of feature extraction. However, the research acknowledged the subjective aspect of visual similarity, pointing out that it does not necessarily match meteorological similarity. To test the feasibility of this strategy, the researchers analyzed meteorological variables such as wind speed and direction across multiple days. Finally, our strategy aims to help forecasters obtain important historical data more rapidly, enhancing the empirical analysis and accuracy of current weather forecasts \cite{16}.

Kang et al. concentrated on building distributed parallel methods for training deep neural network models, especially convolutional neural networks (CNNs), to accelerate the extraction of comparable weather maps, which is critical for effective and fast weather forecasting. They compared the performance of various algorithms in both single and multi-node systems, with numbers of GPUs. It discovered that while employing several GPUs could considerably reduce training time, adding additional nodes in a distributed environment could increase training time owing to network overhead. The conclusion reiterated similar findings, highlighting the efficiency of parallel processing in single-node, multi-GPU configurations for training deep learning models in weather forecasting applications \cite{17}.

Hakii et al. suggested a novel method for predicting weather maps by extracting RGB metaphorical features from atmospheric pressure patterns. This unique technology utilized machine learning to autonomously produce predict weather maps, combining meteorologists' knowledge with observational data. The technique took atmospheric pressure information from current weather maps and converted them to RGB metaphorical gradation maps before employing pix2pix to predict future weather images. The experimental results validated the proposed method's ability to forecast the weather, yielding the possibility of generating predictive weather maps automatically. However, the authors also recognized the need for additional improvement in illustrating pressure patterns and investigating extended forecast capabilities \cite{18}.

Zhang et al. employed deep learning convolutional neural networks and the YOLO V3 models to improve rainfall forecast accuracy by identifying high-altitude, low-pressure vortices on weather maps. The effectiveness of this method was demonstrated by its integration with the MICAPS platform, which enabled forecasters to leverage MICAPS data for real-time database modifications.  While the study did not specify the actual amount of data points utilized in the experiment, it did highlight the enhanced accuracy of rainfall forecasts using this strategy. This finding underlined the relationship between the forecaster's operating system competency and the capacity to access updated data from MICAPS in real-time \cite{19}.

Fang et al. addressed the importance of deep learning for predicting extreme weather occurrences. Due to the unexpected nature and complexity of atmospheric motions, forecasting severe weather is a difficult meteorological challenge. Deep learning approaches, which can learn from huge amounts of data, have proven performance in various tasks. Fang et al. summarized known approaches for extreme weather prediction, by emphasizing employing recurrent neural networks and convolutional neural networks to forecasting, and extracting extreme weather picture data, respectively. This research focused on the widespread and efficient use of deep learning in extreme weather forecasting. It emphasized deep learning's potential to increase forecasting accuracy and lead to the development of new approaches for predicting extreme weather occurrences \cite{20}.

The current study focused on applying deep learning models to detect comparable synoptic weather maps using metrics such as Root Mean Square Error (RMSE) and Structural Similarity Index (SSIM). Since there is still uncertainty about using those metrics, our goal is to discover the most relevant parameters for detecting related weather maps, laying the groundwork for future study in this area.  Our research also utilizes both quantitative and computational methodologies with qualitative observations from experienced meteorologists. By doing so, we want to help understand weather patterns and their consequences, providing thorough insights that combine AI-based technologies with experienced human judgment.

\section{Materials and Methods}
\subsection{Dataset Description and Preprocessing}

The Korea Meteorological Administration (KMA) creates synoptic weather maps for the East Asian area focused on the Korean Peninsula (KP), from surface to vertical level height (Ground Diary, n.d.), as seen in Figure 1. The 500 hPa pressure level, located around 5.5 km above sea level and related to the troposphere's middle layer (Mohanakumar et al., 2008), is significant in these analyses. Analyzing weather patterns at this altitude is very important because it provides insights into atmospheric behavior without being influenced by surface friction or excessive vertical movements. Meteorologists analyze long and short waves in the middle troposphere to identify pressure troughs or ridges and understand the development or decay of weather systems. Therefore, for our experiment, the dataset was downloaded from the Korea Meteorological Administration. It includes the geopotential height (GPH) in meters, representing vertical height from sea level at the 500 hPa pressure altitude plane using solid blue lines with 60-meter intervals. Additionally, temperature in degrees Celsius is indicated by red dashed lines with 5°C intervals. Wind information is conveyed through black wind barbs in knots. Local high and low-pressure areas are marked by a blue "H" and red "L," while local high and low-temperature regions are identified by a red "W" and blue "C," respectively.

\begin{figure}[h!]
	\centering
	\includegraphics[width=0.8\textwidth]{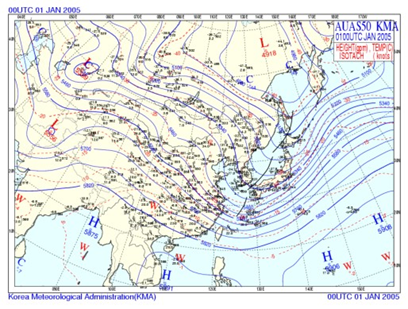}
	\caption{Synoptic weather map by the Korea Meteorological Administration, detailing atmospheric conditions over East Asia}
	\label{fig:fig1}
\end{figure}

In this research, we introduce an image preparation method designed specifically for synoptic weather map improvement. This method uses advanced image alteration techniques to improve the comprehensibility of meteorological data. Drawing a black rectangle over previously defined areas is the first step. This is a practical method of concealing important or superfluous information without compromising the overall integrity of the image. Furthermore, the script employs a sophisticated cropping method that allows for the precise isolation of regions of interest. The cropping settings are scalable, guaranteeing adaptability to various map formats and kinds. Utilizing color masking in the HSV (Hue, Saturation, Value) color space to isolate particular color ranges, especially red and blue tones, exists in the next step. In meteorological contexts, this phase is essential since it improves the visibility of key components such as weather fronts, pressure lines, and temperature gradients. This approach proposes a more precise investigation of weather patterns, which is crucial for improving meteorological research and forecasting. The outcome of the processing is shown in Figure 2.

\begin{figure}[h!]
	\centering
	\includegraphics[width=0.8\textwidth]{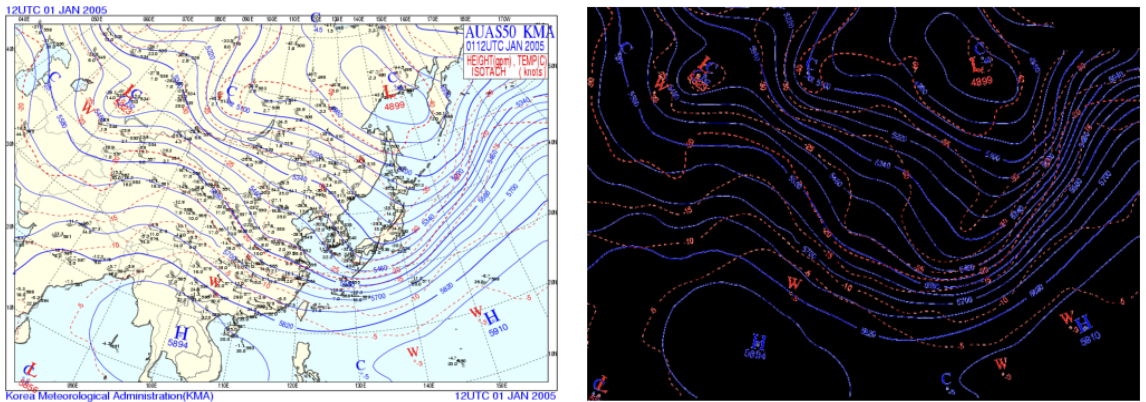}
	\caption{Original Synoptic weather map by the Korea Meteorological Administration (left), and the result of the proposed data preprocessing (right), respectively}
	\label{fig:fig1}
\end{figure}

\subsection{ConvNeXt}
ConvNeXt represents a significant advancement in convolutional neural network (CNN) architectures, distinguishing itself from Transformer-based models like ViT and Swin Transformer. This model is differentiated from using patch embedding or attention, which is a common feature in Transformer architectures, and instead optimizes the traditional CNN approach. It combines traditional CNN architecture with elements inspired by the Swin Transformer's block structures. This hybrid approach allows it to utilize the strengths of both CNNs and Transformers. Additionally, ConvNeXt adopts a training strategy from Swin Transformer, that further enhances its performance, particularly in complex computer vision tasks. With these modifications, the ConvNeXt achieved higher performance compared to that of the Swin Transformer, or VIT \cite{21}.
In the field of meteorological analysis, interpreting synoptic weather maps requires a sophisticated comprehension of both small-scale and large-scale spatial interactions. The complex patterns of isobars, fronts, and symbols call for a model that can identify minute elements and place them in a larger meteorological situation. Conventional CNNs are good at identifying local characteristics like edges and textures, but they can be ineffective in capturing the long-range relationships that are present in weather systems. On the other hand, without substantial pre-training, Vision Transformers (ViTs) may fail to recognize the importance of local spatial hierarchies despite their superior modeling of these global interconnections \cite{22}. It also takes over Transformer models' global receptive field, which enables a comprehensive understanding of weather processes over the whole map. ConvNeXt is guaranteed to be sensitive to both the minute atmospheric characteristics and the complex web of interactions that characterize synoptic weather patterns because of its dual capacity. Furthermore, ConvNeXt's training approach is designed to leverage these architectural advantages with a routine that highlights its two advantages. ConvNeXt is taught to recognize intricate patterns and correlations that are crucial for weather forecasting and research by utilizing current algorithmic advancements. The application of it in our research is strongly supported by its state-of-the-art performance in computer vision benchmarks, which highlights its potential to greatly improve meteorological data interpretation. All things considered, ConvNeXt is an excellent candidate for the complex task of synoptic weather map analysis as it is an ideal combination of inductive biases from both CNNs and ViTs \cite{23}.

\subsection{AutoEncoder}
Due to their unique design and data processing powers, autoencoders—a subset of neural networks—show remarkably high efficacy in identifying similarities across datasets \cite{24}. The encoder, which compresses incoming data into a compressed, lower-dimensional representation, and the decoder, which reconstructs the data from its compressed form, are the two primary components of this architecture. Autoencoders can preserve important characteristics while removing unnecessary information due to this compression of high-dimensional data into a more useful format \cite{25}. As a result, the design not only simplifies data but also highlights its most important characteristics, improving the accuracy with which patterns and similarities may be found. In synoptic meteorology, where labeled datasets are often scarce, analogous weather maps can be found by utilizing autoencoders' built-in feature extraction capabilities. Fundamentally, an autoencoder's ability to learn self-representation relies on its ability to compress high-dimensional meteorological data into a small latent space, capturing significant atmospheric properties like temperature variations, pressure configurations, and hygrometric distributions \cite{26}. Figure 3 depicts the general design of the autoencoder. The encoder portion aims to embed these notable characteristics into a lower dimension’s subspace, while the decoder component tries to recreate the original synoptic maps. This reconstruction requires the preservation of meteorological characteristics required for synoptic analysis. 
Therefore, the autoencoder gains proficiency in identifying subtle relationships between weather maps, which is essential when dealing with novel or never-before-seen meteorological input. Given the autoencoder's capability of creating condensed feature vectors, it is possible to locate nearby weather maps in this latent space and employ closest-neighbor search techniques to find synoptic alternates. This unsupervised approach avoids the tedious process of data annotation, which is highly beneficial for meteorological undertakings. Autoencoders apply the self-representation principle to facilitate the grouping of synoptic maps, which advances the identification and integration of meteorological trends in large-scale historical data archives. It is expected that this method would improve the scope and accuracy of meteorological evaluations, simplifying the process of rapidly identifying significant weather patterns without the need for particular labeling. There are important ramifications for weather forecasting, serving as a beacon for timely and informed decision-making in the face of shifting atmospheric dynamics. 

\begin{figure}[h!]
	\centering
	\includegraphics[width=0.8\textwidth]{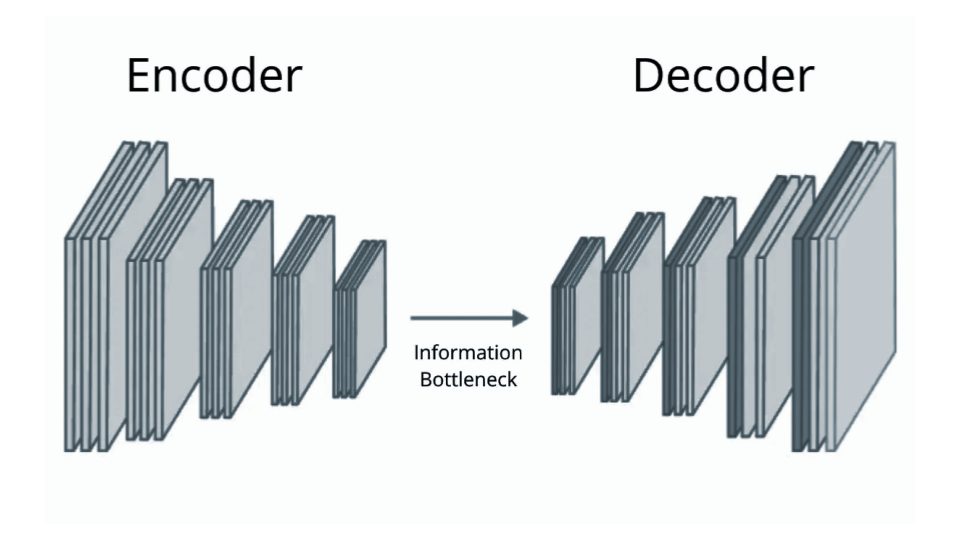}
	\caption{Schematic representation of a Convolutional Autoencoder neural network}
	\label{fig:fig1}
\end{figure}

\subsection{VQ-VAE}
The VQ-VAE method combines classic Variational AutoEncoders (VAEs) with vector quantization approaches. It takes advantage of VAEs' capabilities to produce high-quality, diversified outputs while resolving some limitations through vector quantization. VQ-VAEs have the same encoder-decoder construction as normal VAEs. The encoder compresses the input data into a latent space representation, and the decoder reconstructs the original input from this latent space. The key distinguishing feature of VQ-VAEs is the use of a discrete latent space. The continuous output of the encoder is quantized into a finite set of vectors (a codebook), and the nearest vector in this codebook represents each input. This quantization phase promotes more stable training and yields higher-quality outcomes, especially in situations where data is naturally classified into discrete categories. Three separate loss functions used to train the VQ-VQE are reconstruction loss, codebook loss, and commitment loss, respectively. The reconstruction loss denotes how well the decoder can reconstruct the original input from the quantized latent representations. A lower reconstruction loss indicates better performance. The role of the codebook loss is to ensure that the codebook vectors effectively represent the data, a loss is applied to make sure that each vector in the codebook is as close as possible to the encoder's output. For the last, the commitment loss encourages the encoder to produce outputs close to the codebook vectors, ensuring stability and consistency in the mapping \cite{27}. The core strength of a VQ-VAE lies in its ability to compress complex, high-dimensional meteorological data into a discrete and compact latent space. It excels in capturing critical atmospheric features like pressure systems, temperature gradients, and moisture distributions. In a VQ-VAE, the encoder reduces these key meteorological characteristics to a lower-dimensional subspace, while the decoder attempts to recover the original synoptic maps from this compressed representation. This reconstruction procedure is critical because it assures the preservation of essential meteorological properties required for effective synoptic analysis. As a result, the VQ-VAE improves its ability to recognize complex correlations between multiple weather maps, which is critical for studying fresh or previously unheard meteorological data. The discrete character of the latent space in VQ-VAEs, which is represented by a finite collection of vectors or 'codebook', enables the effective localization of comparable weather maps. Synoptic analogs may be easily identified by using nearest-neighbor search algorithms in this quantized latent space.

\section{Result}
Our research evaluated an array of state-of-the-art pre-trained deep learning models to ascertain their efficacy in identifying synoptic weather map patterns. In the first phase, we utilized models famous for their image classification performance, such as VGG16 and VGG19 [28]. We also examined models like as Xception and InceptionV3, which have more sophisticated and efficient architectures, as well as ResNet50, which is well-known for its residual learning framework, which allows for the training of deeper networks [29][30][31]. We also used EfficientNet models, including EfficientNetB0 and EfficientNetB1, which are known for their scalable design and balance of accuracy and efficiency [32]. For the most recent phases, EfficientNetV2B0 and EfficientNetV2B1, which enhance training speed and parameter efficiency, were used [33]. Most importantly, we included the ConvNeXtTiny and ConvNeXtBase models in our research. ConvNeXt's hybrid design, which drew on the capabilities of both CNNs and Transformers, served as a focal point for comparison due to its current architecture and cutting-edge performance on a range of computer vision applications. We then used our dataset to train an autoencoder that we had previously deployed. Lastly, we used three evaluation metrics: cosine similarity, structural similarity index (SSIM), and root mean square error (RMSE). Our goal was to determine which metric best matches a meteorologist's perspective when searching for comparable synoptic weather maps.

\subsection{Results from Convolutional Autoencoder}
The input data for the experiment is the 500 hPa synoptic weather map at 00 UTC on December 19, 2022. In the examination of five historical weather maps generated by the proposed autoencoder model results, notable patterns emerge. The result yielded that based on the metrics about SSIM, and RMSE, even though the given input is the December data, they described the result from September, and even from the summer. It could be inferred that those metrics failed to figure out the similar ones effectively, which means that it might not be suitable to use those metrics for this task.
However, the results from cosine similarity showed much more ideal results compared to other metrics used above and those results are described in Figure 4. Firstly, closed low-pressure systems, as previously mentioned, are consistently distributed in similar geographic locations across these maps. Furthermore, the positions of troughs and ridges resulting from Rossby wave dynamics exhibit remarkable similarities. One noteworthy observation is the presence of a trough around the KP (Key Point) region, resulting in a distinct narrow effect on geopotential height (GPH) and temperature lines. When assessing these results on a ranked scale from 1 to 5, the top-ranking result, designated as number 1, stands out for accurately depicting both the number and location of closed low-pressure systems with cold cores, as well as the position of the COL (Cut-Off Low) point. Furthermore, the result's trough east of KP is in good agreement with the temperature and wind pattern distribution seen in the input weather map. Two troughs are seen around the KP region in the second-ranked result (number 2), which leads to a more noticeable pressure gradient and, in turn, more violent winds. Wider gaps in the GPH and temperature lines surrounding KP are observed in the third-ranked result (number 3), which is explained by a trough that lies farther east. The fourth result shows that a brief wave trough is moving eastward and is about to strike KP. The north-south temperature gradient on KP looks to be a little weaker in this finding. Lastly, finding number 5 differs from the others in that it shows that the trough is getting closer to KP rather than having crossed it yet. These evaluations collectively indicate that while variations exist among the ranked results, the model is adept at identifying historical weather maps that bear striking similarities to the current input, offering valuable insights into the dynamics of weather patterns. Furthermore, with this analysis, we could figure out that the most suitable metric for discovering a similar synoptic weather map is the cosine similarity, as shown in Table 1.

\begin{figure}[h!]
	\centering
	\includegraphics[width=0.8\textwidth]{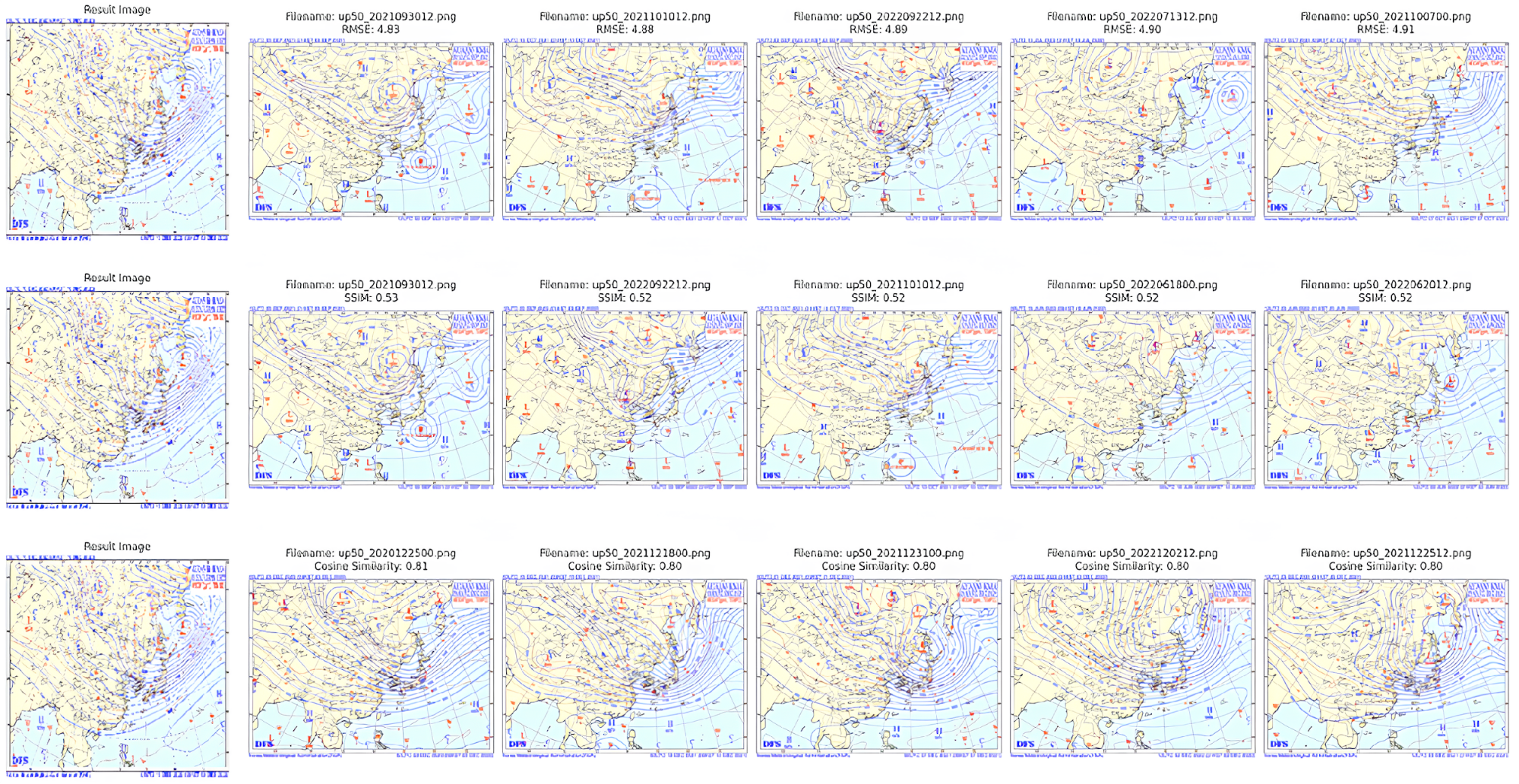}
	\caption{A set of synoptic weather maps displays a series of synoptic weather maps generated by an Autoencoder model, assessed using three metrics: RMSE (top row), SSIM (middle row), and Cosine Similarity (bottom row), providing a multi-faceted evaluation of model accuracy}
	\label{fig:fig1}
\end{figure}

\begin{table}[h]
\centering
\caption{Comparative analysis of image quality metrics across five test samples, showcasing root mean square error (RMSE), structural similarity index (SSIM), and cosine similarity values.}
\begin{tabular}{@{}lccccc@{}}
\toprule
\textbf{Metric} & \textbf{\#1} & \textbf{\#2} & \textbf{\#3} & \textbf{\#4} & \textbf{\#5} \\ \midrule
\textbf{RMSE}   & 4.83         & 4.88         & 4.89         & 4.90         & 4.91         \\
\textbf{SSIM}   & 0.53         & 0.52         & 0.52         & 0.52         & 0.52         \\
\textbf{Cosine} & 0.81         & 0.80         & 0.80         & 0.80         & 0.80         \\
\textbf{Similarity} &           &             &             &             &              \\ \bottomrule
\end{tabular}
\end{table}

\subsection{Results from VQ-VAE}
We next used the VQ-VQE to the same dataset for comparison, as the Convolutional Autoencoder (CAE) is not a novel method. The general approach is very similar since, as previously said, this method also includes an encoder and a decoder.  Initially, we were able to locate the rebuilt pictures after VQ-VQE training. Following data processing by the VQ-VAE's encoder and decoder network, Figure 5's reconstructed image on the right aims to replicate these properties. The visual examination indicates that the rebuilt map is identical to the original in terms of general arrangement and meteorological data placement. The isobars appear to preserve their relative shapes and trajectories, and the high and low-pressure systems are evident in positions consistent with the original map. However, because the VQ-VAE compresses and reconstructs the input data into a lower-dimensional latent representation before creating the output, there may be minor changes in the smoothness of the lines or subtle differences in the details.

Figure 6 depicts a comparison between an original synoptic weather map and its corresponding encoded representation, termed here as "Code." With distinctive meteorological features like isobars—curved lines that indicate areas of equal atmospheric pressure—and symbols for high-pressure (designated with an 'H') and low-pressure (marked with an 'L') systems, the left panel, titled "Original," displays a comprehensive synoptic weather map. On the right, the panel labeled "Code" displays the encoded version of the weather map as processed by a VQ-VAE model. This representation is shown as a grid of colored squares, with each color denoting a unique vector found in the codebook of the model. The variety of hues points to a rich encoding of the properties of the map into a separate, lower-dimensional latent space. This encoded form abstracts the continuous data from the weather map into a format that may be used by the model for pattern recognition, similarity searches, or reconstruction of the input data.

\begin{figure}[h!]
	\centering
	\includegraphics[width=0.8\textwidth]{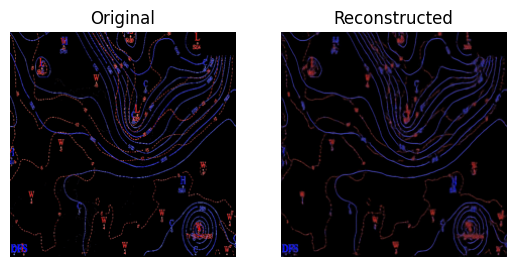}
	\caption{Comparison of Original and VQ-VAE Reconstructed Synoptic Weather Maps}
	\label{fig:fig1}
\end{figure}

\begin{figure}[h!]
	\centering
	\includegraphics[width=0.8\textwidth]{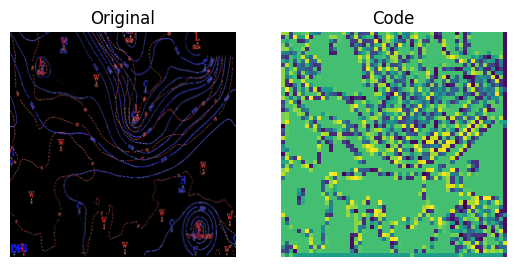}
	\caption{Synoptic Weather Map and Its Discrete Latent Encoding by VQ-VAE}
	\label{fig:fig1}
\end{figure}

A series of synoptic weather maps are shown in Figure 7, with the first map designated as the "Anchor" and the remaining maps, labeled 1 through 5, being determined to be comparable to the anchor map by a high cosine similarity score of 0.98. Even with the high similarity ratings, there are still observable discrepancies between the maps regarding synoptic aspects such as pressure system location and form, isobar configuration, and frontal boundary depiction. The subtleties and complexity that a meteorologist would take into account in a real-world situation are not sufficiently reflected in the parallels that the quantitative score suggests. This mismatch demonstrates the limitations of using cosine similarity scores alone to evaluate synoptic weather map similarity. The ratings may not fully convey the expert judgment needed for proper meteorological analysis, which frequently requires a more nuanced and context-specific interpretation of the data. The model's evaluation cannot entirely replace the qualitative aspects required for operational forecasting, which include the specific configuration and interplay of meteorological variables.

\begin{figure}[h!]
	\centering
	\includegraphics[width=0.8\textwidth]{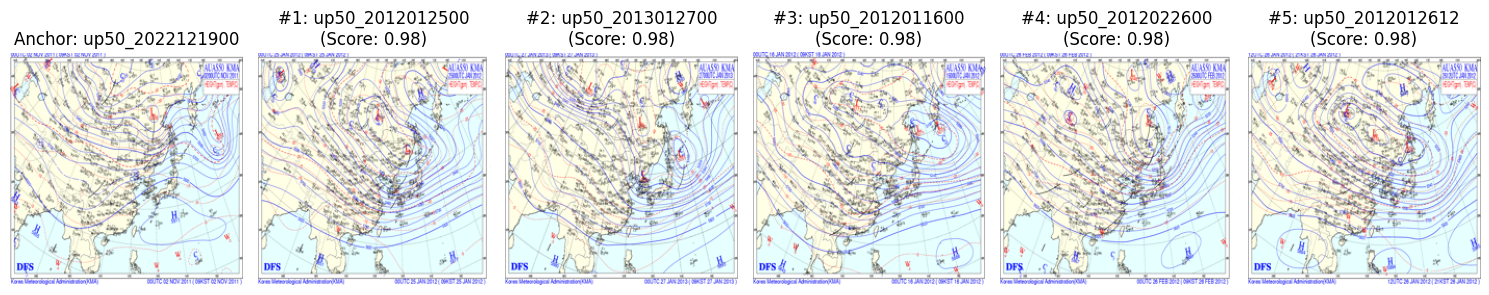}
	\caption{A set of synoptic weather maps displays a series of synoptic weather maps generated by VQ-VAE}
	\label{fig:fig1}
\end{figure}

Proposed autoencoders can represent data in a continuous latent space (Zhao et al., 2018). This indicates that the encoder creates a point in a multi-dimensional space with true coordinates for each input (a synoptic weather map in this example), enabling endlessly fine gradations between points. The model can accurately represent the subtle and gradual shifts in the input data due to the continuous character of this space. In the context of weather maps, this might imply faithfully depicting the slow and continuous changes that occur in meteorological phenomena such as moisture levels, temperature swings, and pressure gradients, as opposed to abrupt shifts (Cassano et al., 2006). A VQ-VAE, on the other hand, maps inputs to a discrete collection of latent space points (Razavi et al., 2019). This is comparable to allocating each input to the closest category out of a limited number of predetermined categories (the codebook vectors). This may not always be the greatest option for data that shows a high degree of continuity, even if it may be quite useful for some types of data, particularly those with distinct, well-defined categories. The depiction of continuous atmospheric variables such as pressure systems, frontal borders, and isobars is a representational problem for Vector Quantized-Variational AutoEncoders (VQ-VAEs) in synoptic weather maps. Conventional autoencoders are good at capturing the delicate gradations and smooth transitions found in meteorological data, including the gentle shifts that identify the borders of air masses and the smooth flow of isobars that indicate equal pressure. These autoencoders have continuous latent spaces. These features, often existing on a continuum in the real atmosphere, are better represented without the abrupt categorization imposed by VQ-VAEs. VQ-VAEs, though beneficial for their stability and categorical clarity, may falter with weather data where the gradience of atmospheric conditions and the integrity of detailed features are paramount. A separated depiction that may miss the fluidity seen in synoptic weather maps might result from the process of quantizing data into a discrete collection, which can unintentionally leave out important features and break the continuity of weather patterns. This restriction implies that continuous latent spaces may be preferred in situations where faithful preservation of continuous variables is essential.

\subsection{Results from Imagenet-based Pretraiend Models}
We used an experimental approach in this work, using pretrained deep learning models for analytical comparison—a methodology that has been used in comparable studies. Our method was applying these advanced models to a dataset that had been specifically preprocessed. First, five well-known models—VGG16, VGG19, Xception, InceptionV3, and ResNet50—were chosen for this task. The choice of ImageNet, despite its stark contrast to the unique characteristics of synoptic weather maps, was strategic. Our dataset, after undergoing preprocessing, primarily featured line representations of images. This conversion led us to suppose that the information acquired from ImageNet's massive collection of photos would still be practical and valuable in this case. The majority of the pictures in our sample were represented as lines following preprocessing. This conversion led us to believe that the insights gathered from ImageNet's large picture collection will remain relevant and useful.

Furthermore, we took inspiration from prior studies that demonstrated the importance of cosine similarity as an analytical measure. As a result, we narrowed our emphasis to one specific parameter, hoping to draw the most significant comparisons across model results. Our investigation produced intriguing findings. For example, the VGG16 and VGG19 models produced findings in March and April. This discovery was intriguing, but it did not fully meet our expectations because the target photographs were from December. Xception, InceptionV3, and ResNet, on the other hand, were more closely matched with the desired timeframe, with outputs largely occurring in November and October. This result indicated a more sophisticated knowledge of the models' capabilities and applicability to different types of datasets, particularly when considering the temporal characteristics of the data they were trained to comprehend. Figures 8 to 12 display extensive visual representations based on cosine similarity scores, while Table 2 provides accurate numerical data.

\begin{table}[h]
\centering
\caption{Comparative analysis of image quality metrics across five test samples, showcasing cosine similarity values.}
\begin{tabular}{@{}lccccc@{}}
\toprule
\textbf{Metric} & \textbf{\#1} & \textbf{\#2} & \textbf{\#3} & \textbf{\#4} & \textbf{\#5} \\ \midrule
VGG16           & 1.00         & 0.82         & 0.81         & 0.76         & 0.78         \\
VGG19           & 1.00         & 0.83         & 0.81         & 0.76         & 0.75         \\ 
Xception        & 1.00         & 0.87         & 0.82         & 0.81         & 0.80         \\
InceptionV3     & 1.00         & 0.80         & 0.78         & 0.77         & 0.77         \\
ResNet50        & 1.00         & 0.84         & 0.81         & 0.77         & 0.76         \\ \bottomrule
\end{tabular}
\end{table}

\begin{figure}[h!]
	\centering
	\includegraphics[width=0.8\textwidth]{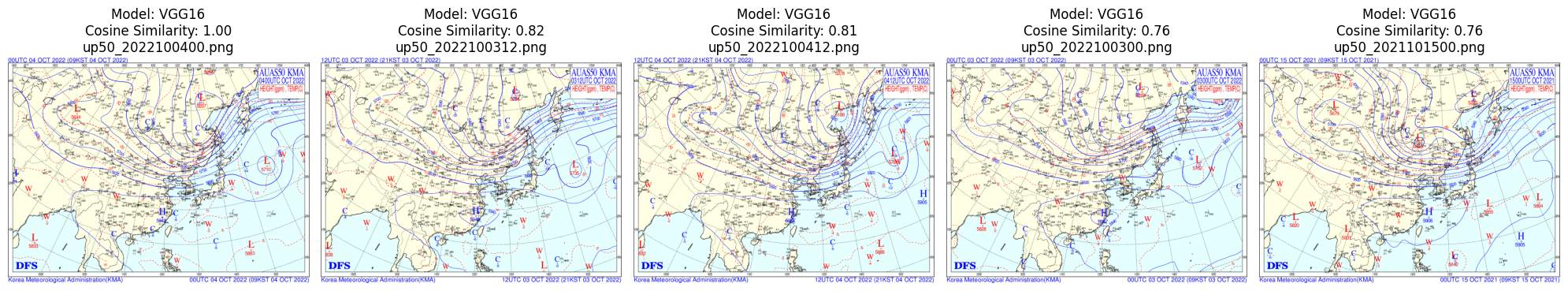}
	\caption{A set of synoptic weather maps displays a series of synoptic weather maps generated by VGG16}
	\label{fig:fig1}
\end{figure}

\begin{figure}[h!]
	\centering
	\includegraphics[width=0.8\textwidth]{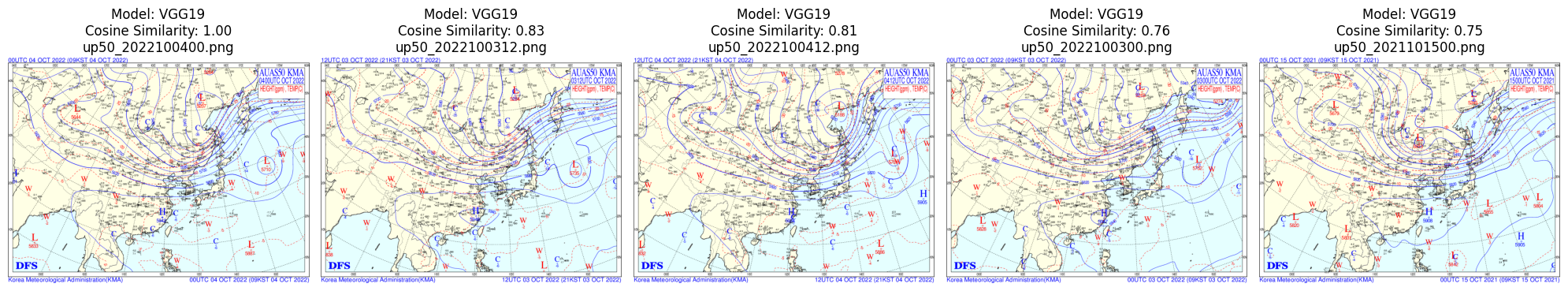}
	\caption{A set of synoptic weather maps displays a series of synoptic weather maps generated by VGG19}
	\label{fig:fig1}
\end{figure}

\begin{figure}[h!]
	\centering
	\includegraphics[width=0.8\textwidth]{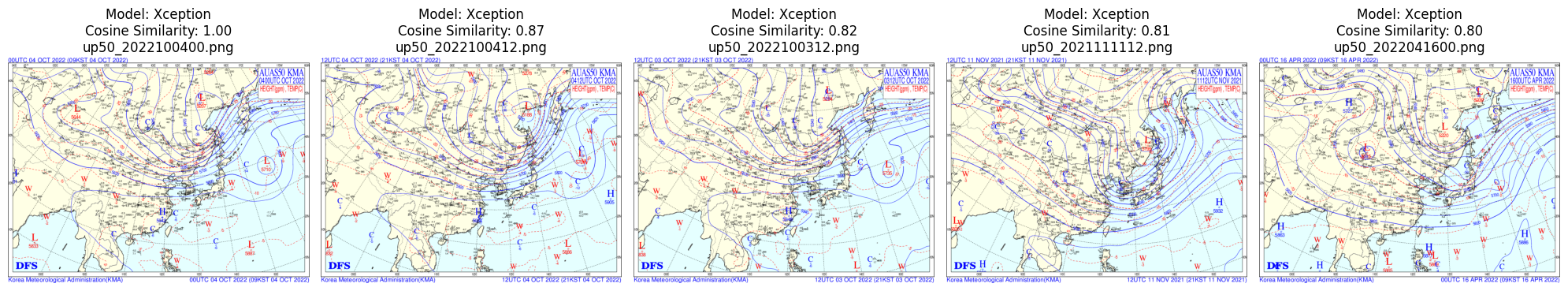}
	\caption{A set of synoptic weather maps displays a series of synoptic weather maps generated by Xception}
	\label{fig:fig1}
\end{figure}

\begin{figure}[h!]
	\centering
	\includegraphics[width=0.8\textwidth]{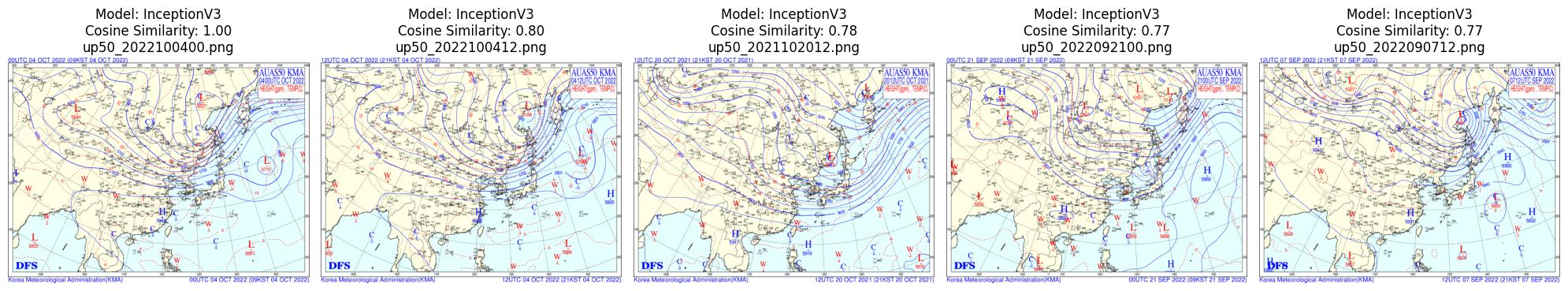}
	\caption{A set of synoptic weather maps displays a series of synoptic weather maps generated by InceptionV3}
	\label{fig:fig1}
\end{figure}

\begin{figure}[h!]
	\centering
	\includegraphics[width=0.8\textwidth]{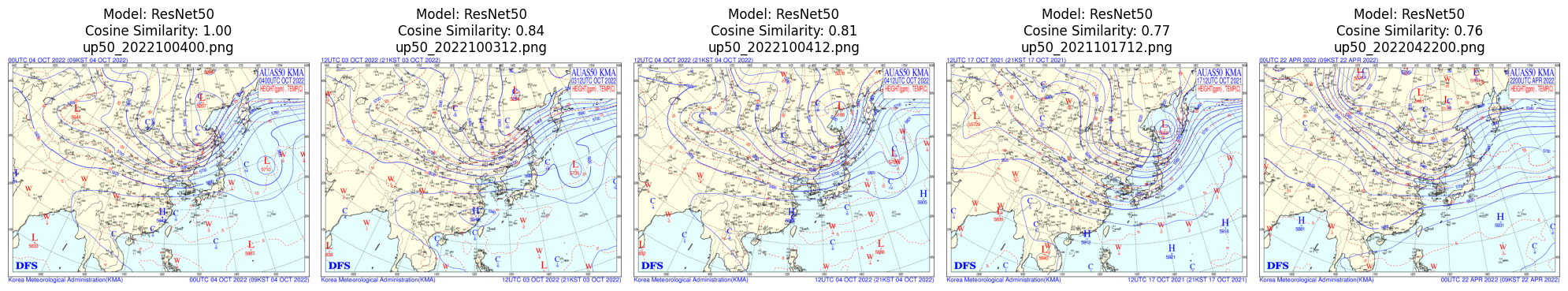}
	\caption{A set of synoptic weather maps displays a series of synoptic weather maps generated by ResNet50}
	\label{fig:fig1}
\end{figure}

We have included several more recent models, such as EfficientNetB0, EfficinetNetB1, EfficientNetV2B0, EfficientNetV2B1, ConvNeXtTiny, and ConvNeXBase, in our thorough investigation of pre-trained deep learning models. Figures 13 to 18  reveal that these models produced datasets primarily from October and November, which is intriguing and suggests that they performed better than the models VGG16 and VGG19. Their more recent development may have contributed to their better performance, indicating that more current models may produce more accurate findings. The consistency of the outcomes across all models was another intriguing feature of our research. With a cosine similarity score of 1.0, every model determined that the same image was the most similar to our goal. This striking uniformity highlights the important impact of using the ImageNet dataset for training. It suggests that although these models had different architectures, their initial training on ImageNet imposed a feature on their processing that allowed them to converge on a single, very similar output. This discovery indicates the possible influence of training datasets on the performance and behavior of deep learning models in addition to highlighting the robustness of these models. Furthermore, Table 3 displays the exact results obtained from the studies.

\begin{figure}[h!]
	\centering
	\includegraphics[width=0.8\textwidth]{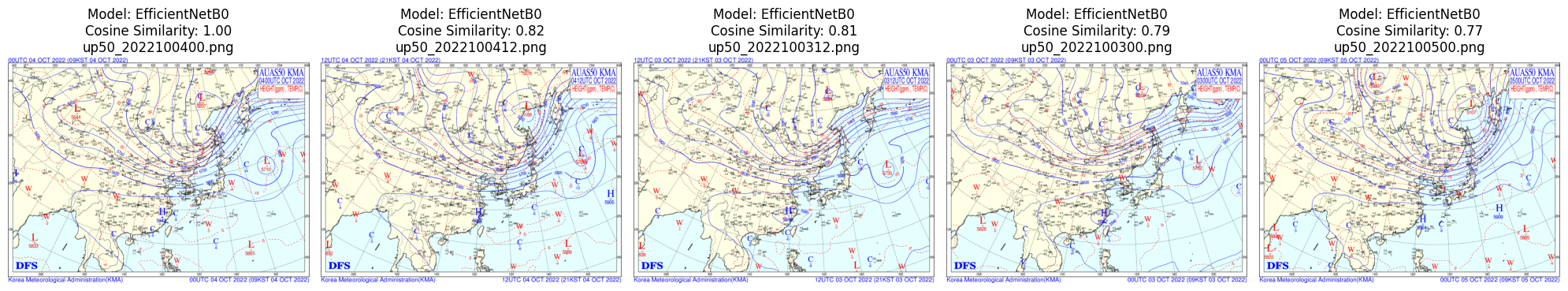}
	\caption{A set of synoptic weather maps displays a series of synoptic weather maps generated by EfficientNetB0}
	\label{fig:fig1}
\end{figure}

\begin{figure}[h!]
	\centering
	\includegraphics[width=0.8\textwidth]{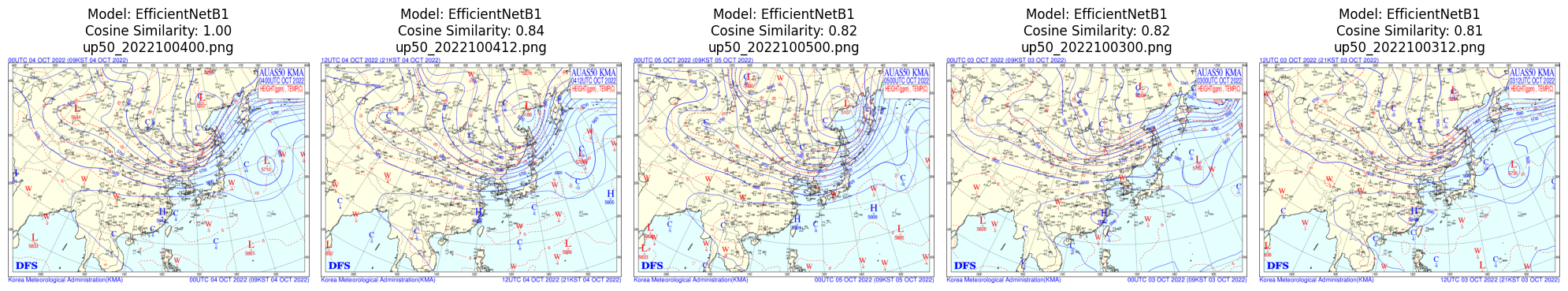}
	\caption{A set of synoptic weather maps displays a series of synoptic weather maps generated by EfficientNetB1}
	\label{fig:fig1}
\end{figure}

\begin{figure}[h!]
	\centering
	\includegraphics[width=0.8\textwidth]{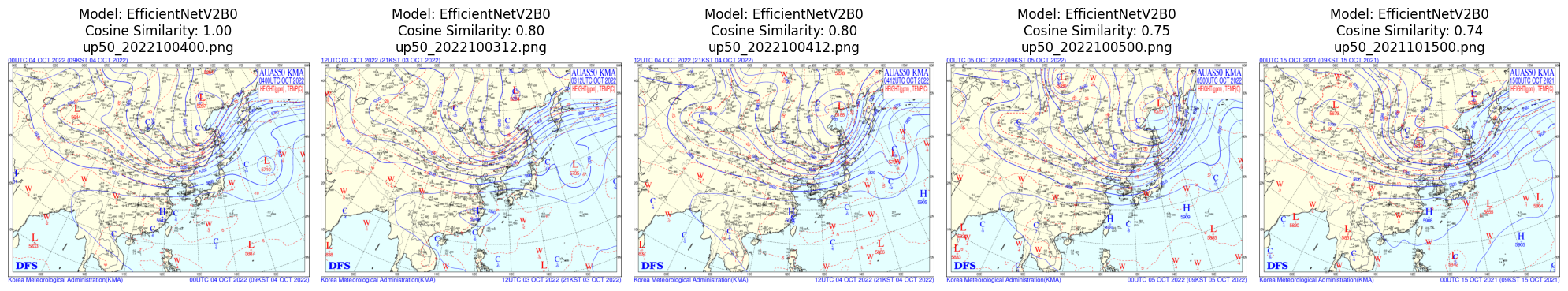}
	\caption{A set of synoptic weather maps displays a series of synoptic weather maps generated by EfficientNetV2B0}
	\label{fig:fig1}
\end{figure}

\begin{figure}[h!]
	\centering
	\includegraphics[width=0.8\textwidth]{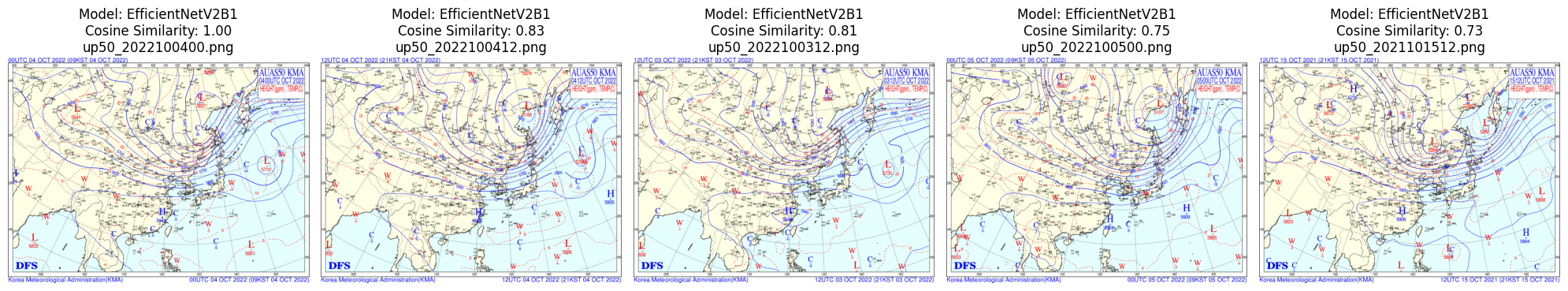}
	\caption{A set of synoptic weather maps displays a series of synoptic weather maps generated by EfficientNetV2B1}
	\label{fig:fig1}
\end{figure}

\begin{figure}[h!]
	\centering
	\includegraphics[width=0.8\textwidth]{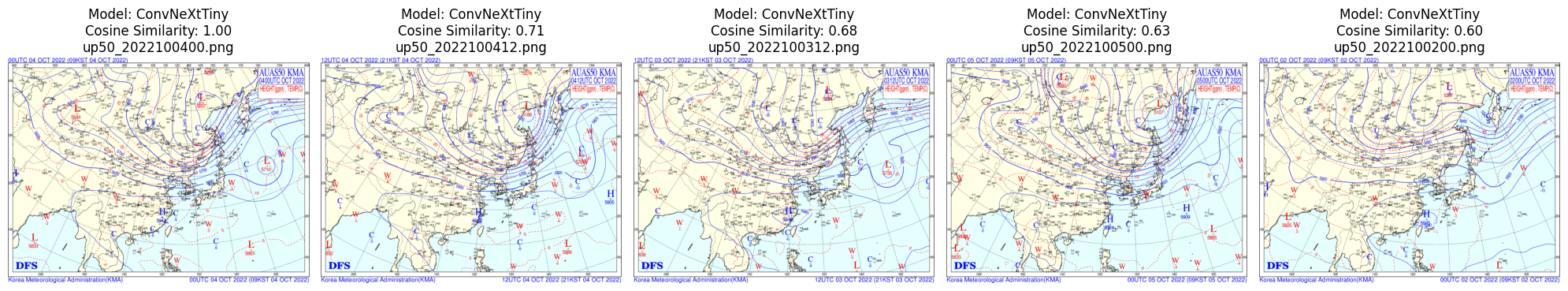}
	\caption{A set of synoptic weather maps displays a series of synoptic weather maps generated by ConvNeXtTiny}
	\label{fig:fig1}
\end{figure}

\begin{figure}[h!]
	\centering
	\includegraphics[width=0.8\textwidth]{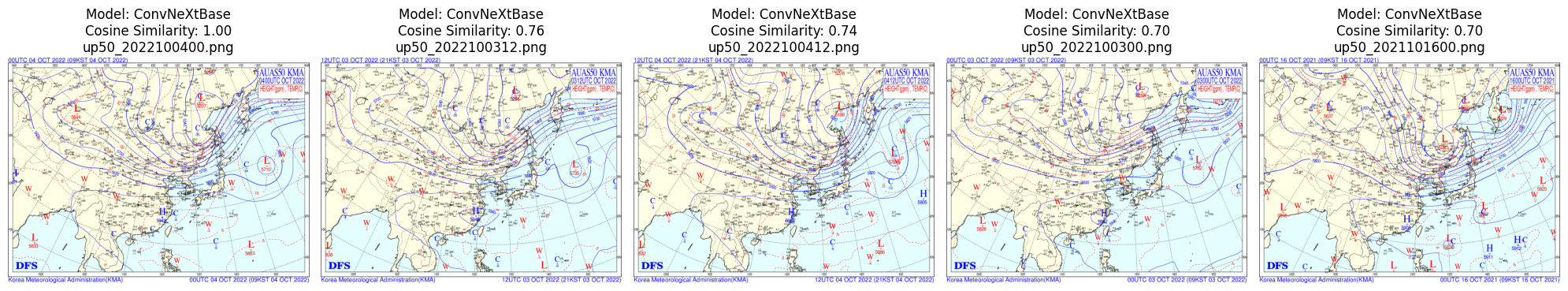}
	\caption{A set of synoptic weather maps displays a series of synoptic weather maps generated by ConvNeXtBase}
	\label{fig:fig1}
\end{figure}

\begin{table}[h]
\centering
\caption{Comparative analysis of image quality metrics across five test samples, showcasing cosine similarity values with different pretrained models.}
\begin{tabular}{@{}lccccc@{}}
\toprule
\textbf{Metric}       & \textbf{\#1} & \textbf{\#2} & \textbf{\#3} & \textbf{\#4} & \textbf{\#5} \\ \midrule
EfficientNetB0        & 1.00         & 0.82         & 0.81         & 0.77         & 0.69         \\
EfficientNetB1        & 1.00         & 0.84         & 0.82         & 0.81         & 0.69         \\
EfficientNetV2B0      & 1.00         & 0.80         & 0.80         & 0.74         & 0.64         \\
EfficientNetV2B1      & 1.00         & 0.83         & 0.81         & 0.73         & 0.61         \\
ConvNeXtTiny          & 1.00         & 0.71         & 0.68         & 0.70         & 0.53         \\
ConvNeXtBase          & 1.00         & 0.76         & 0.74         & 0.60         & 0.64         \\ \bottomrule
\end{tabular}
\end{table}

\section{Discussion}
The investigation highlights the potential use of deep learning algorithms to recognize comparable synoptic weather maps. It introduces a novel data processing approach designed for synoptic weather map difficulties. To find the most similar maps, a variety of deep learning models were used, including cutting-edge models such as ConvNext. By combining these innovative algorithms and contrasting multiple approaches, this study significantly improved on past research. Incorporating meteorologists' qualitative remarks was particularly important in confirming the findings of the deep learning models. However, this work recognizes a significant limitation: relying entirely on deep learning models imposes some limits, especially because the differential in scores produced by these models is not always evident. This uncertainty demands the cooperation of meteorologists to have a more thorough knowledge and accurate forecast. As a result, the suggested technique should be considered a decision support system rather than a stand-alone solution. Our model-based decision support system will easily interact with existing meteorological tools, providing real-time data analysis and suggestions. It will use sophisticated analytics to quickly detect patterns and connections in weather data, giving meteorologists a better grasp of synoptic weather maps. Future studies will focus on improving this decision support system's flexibility, user experience, and interaction with a larger set of meteorological tools.

\section{Conclusion}
The objective of this paper was to use synoptic weather map analysis—a vital meteorological technique for depicting comprehensive weather conditions over large areas—to bridge the gap between the weather prediction model and real forecast operations. Detailed techniques are needed to provide accurate and dependable weather forecasts due to the distinct manner in which synoptic maps depict atmospheric conditions. As a result, we suggested an Autoencoder-based technique that can recognize past weather maps that closely match the situation as it is right now. This model is a technical development in meteorology that aims to improve forecasting by fusing cutting-edge technological capabilities with conventional meteorological information. We also experimented with a Vector Quantized-Variational AutoEncoder (VQ-VAE) to investigate the use of deep learning to synoptic weather map processing. Our research, however, showed that the VQ-VAE did not perform as well as a conventional convolutional autoencoder. This may be explained by the inherent qualities of synoptic weather maps, which are full of smooth and constant changes in meteorological components. The discrete latent space of the VQ-VAE does not capture these continuous properties as well. Several pre-trained deep learning models were used; initially, the focus was on models like VGG16, VGG19, Xception, InceptionV3, and ResNet50 that were trained on the diverse ImageNet dataset. It's interesting to note that the study showed that certain models fit more closely with the targeted schedule than others. This led to the conclusion that although these pre-trained models show impressive performance in a variety of settings, they may not be as effective in specialized tasks such as recognizing identical synoptic weather images.

Our analysis was also extended to incorporate more recent models, such as EfficientNet and ConvNeXt. Even with their enhanced powers, these models also highlighted an important discovery: their consistency in recognizing similar images, probably because of their initial training on ImageNet, indicated a constraint in their ability to adapt to datasets that differ greatly from ImageNet, like synoptic weather maps. Our study also shows the potential of deep learning methods for finding similar synoptic weather maps and provides a baseline for applying deep learning models to the synoptic weather map, especially estimating the metric, data preprocessing method, and comparing the various models. The incorporation of qualitative evaluations from meteorologists confirmed the outcomes of the deep learning models, emphasizing the need for combining computer techniques with knowledgeable human opinion for thorough analysis. Our work proposes a novel data pretreatment technique and relevant criteria for synoptic weather map analysis, representing a significant step forward in meteorological forecasting utilizing artificial intelligence. With weather growing increasingly unpredictable, this gadget will assist meteorologists by delivering real-time data. Our study sheds new light on the ever-changing difficulties of meteorology by connecting numerical precision with real-world applicability.

\bibliographystyle{unsrtnat}
\bibliography{template}

\begin{thebibliography}{27}
\providecommand{\natexlab}[1]{#1}
\providecommand{\url}[1]{\texttt{#1}}
\expandafter\ifx\csname urlstyle\endcsname\relax
  \providecommand{\doi}[1]{doi: #1}\else
  \providecommand{\doi}{doi: \begingroup \urlstyle{rm}\Url}\fi

\bibitem[Karl and Easterling(1999)]{1}
Thomas~R Karl and David~R Easterling.
\newblock Climate extremes: selected review and future research directions.
\newblock \emph{Climatic change}, 42\penalty0 (1):\penalty0 309--325, 1999.

\bibitem[Agyekum et~al.(2022)Agyekum, Antwi-Agyei, and Dougill]{2}
Thomas~Peprah Agyekum, Philip Antwi-Agyei, and Andrew~J Dougill.
\newblock The contribution of weather forecast information to agriculture, water, and energy sectors in east and west africa: A systematic review.
\newblock \emph{Frontiers in Environmental Science}, 10:\penalty0 935696, 2022.

\bibitem[Panteli and Mancarella(2015)]{3}
Mathaios Panteli and Pierluigi Mancarella.
\newblock Influence of extreme weather and climate change on the resilience of power systems: Impacts and possible mitigation strategies.
\newblock \emph{Electric Power Systems Research}, 127:\penalty0 259--270, 2015.

\bibitem[Barry and Chorley(2009)]{4}
Roger~G Barry and Richard~J Chorley.
\newblock \emph{Atmosphere, weather and climate}.
\newblock Routledge, 2009.

\bibitem[Pathak et~al.(2022)Pathak, Subramanian, Harrington, Raja, Chattopadhyay, Mardani, Kurth, Hall, Li, Azizzadenesheli, et~al.]{5}
Jaideep Pathak, Shashank Subramanian, Peter Harrington, Sanjeev Raja, Ashesh Chattopadhyay, Morteza Mardani, Thorsten Kurth, David Hall, Zongyi Li, Kamyar Azizzadenesheli, et~al.
\newblock Fourcastnet: A global data-driven high-resolution weather model using adaptive fourier neural operators.
\newblock \emph{arXiv preprint arXiv:2202.11214}, 2022.

\bibitem[Nguyen et~al.(2023)Nguyen, Brandstetter, Kapoor, Gupta, and Grover]{6}
Tung Nguyen, Johannes Brandstetter, Ashish Kapoor, Jayesh~K Gupta, and Aditya Grover.
\newblock Climax: A foundation model for weather and climate.
\newblock \emph{arXiv preprint arXiv:2301.10343}, 2023.

\bibitem[Bi et~al.(2023)Bi, Xie, Zhang, Chen, Gu, and Tian]{7}
Kaifeng Bi, Lingxi Xie, Hengheng Zhang, Xin Chen, Xiaotao Gu, and Qi~Tian.
\newblock Accurate medium-range global weather forecasting with 3d neural networks.
\newblock \emph{Nature}, 619\penalty0 (7970):\penalty0 533--538, 2023.

\bibitem[Lam et~al.(2023)Lam, Sanchez-Gonzalez, Willson, Wirnsberger, Fortunato, Alet, Ravuri, Ewalds, Eaton-Rosen, Hu, et~al.]{8}
Remi Lam, Alvaro Sanchez-Gonzalez, Matthew Willson, Peter Wirnsberger, Meire Fortunato, Ferran Alet, Suman Ravuri, Timo Ewalds, Zach Eaton-Rosen, Weihua Hu, et~al.
\newblock Learning skillful medium-range global weather forecasting.
\newblock \emph{Science}, 382\penalty0 (6677):\penalty0 1416--1421, 2023.

\bibitem[Chen et~al.(2023)Chen, Han, Gong, Bai, Ling, Luo, Chen, Ma, Zhang, Su, et~al.]{9}
Kang Chen, Tao Han, Junchao Gong, Lei Bai, Fenghua Ling, Jing-Jia Luo, Xi~Chen, Leiming Ma, Tianning Zhang, Rui Su, et~al.
\newblock Fengwu: Pushing the skillful global medium-range weather forecast beyond 10 days lead.
\newblock \emph{arXiv preprint arXiv:2304.02948}, 2023.

\bibitem[Cheon et~al.(2024)Cheon, Choi, Kang, Choi, Lee, and Kang]{10}
Minjong Cheon, Yo-Hwan Choi, Seon-Yu Kang, Yumi Choi, Jeong-Gil Lee, and Daehyun Kang.
\newblock Karina: An efficient deep learning model for global weather forecast.
\newblock \emph{arXiv preprint arXiv:2403.10555}, 2024.

\bibitem[Tarek et~al.(2020)Tarek, Brissette, and Arsenault]{11}
Mostafa Tarek, Fran{\c{c}}ois~P Brissette, and Richard Arsenault.
\newblock Evaluation of the era5 reanalysis as a potential reference dataset for hydrological modelling over north america.
\newblock \emph{Hydrology and Earth System Sciences}, 24\penalty0 (5):\penalty0 2527--2544, 2020.

\bibitem[Lagerquist et~al.(2019)Lagerquist, McGovern, and Gagne~II]{12}
Ryan Lagerquist, Amy McGovern, and David~John Gagne~II.
\newblock Deep learning for spatially explicit prediction of synoptic-scale fronts.
\newblock \emph{Weather and Forecasting}, 34\penalty0 (4):\penalty0 1137--1160, 2019.

\bibitem[Wang et~al.(2006)Wang, Ding, and Sikka]{13}
Bin Wang, Yihui Ding, and DR~Sikka.
\newblock Synoptic systems and weather.
\newblock \emph{The asian monsoon}, pages 131--201, 2006.

\bibitem[Qian et~al.(2021)Qian, Du, and Ai]{14}
Weihong Qian, Jun Du, and Yang Ai.
\newblock A review: Anomaly-based versus full-field-based weather analysis and forecasting.
\newblock \emph{Bulletin of the American Meteorological Society}, 102\penalty0 (4):\penalty0 E849--E870, 2021.

\bibitem[Galen et~al.(2022)Galen, Hartogensis, Benedict, and Steeneveld]{15}
Lars~van Galen, Oscar Hartogensis, Imme Benedict, and Gert-Jan Steeneveld.
\newblock Teaching a weather forecasting class in the 2020s.
\newblock \emph{Bulletin of the American Meteorological Society}, 103\penalty0 (2):\penalty0 E248--E265, 2022.

\bibitem[Ahn et~al.(2023)Ahn, Lee, Ko, Kim, Han, and Seok]{16}
H.~Ahn, S.~Lee, H.~Ko, M.~Kim, S.W. Han, and J.~Seok.
\newblock Searching similar weather maps using convolutional autoencoder and satellite images.
\newblock \emph{ICT Express}, 9\penalty0 (1):\penalty0 69--75, 2023.
\newblock \doi{10.1016/j.icte.2022.03.013}.

\bibitem[Kang et~al.(2018)Kang, Jeong, and Jeong]{17}
B.~Kang, J.H. Jeong, and C.~Jeong.
\newblock Distributed parallel deep learning for fast extraction of similar weather map.
\newblock In \emph{Proceedings of the IEEE Region 10 Conference (TENCON)}, pages 1426--1429, 2018.
\newblock \doi{10.1109/TENCON.2018.8650452}.

\bibitem[Hakii et~al.(2021)Hakii, Shimada, Nakanishi, Okada, Matsuda, Onishi, and Takahashi]{18}
T.~Hakii, K.~Shimada, T.~Nakanishi, R.~Okada, K.~Matsuda, R.~Onishi, and K.~Takahashi.
\newblock Weather map prediction using rgb metaphorical feature extraction for atmospheric pressure patterns.
\newblock In \emph{Proceedings of the 19th IEEE/ACIS International Conference on Computer and Information Science (ICIS)}, pages 22--28, 2021.
\newblock \doi{10.1109/ICIS51859.2021.9462191}.

\bibitem[Zhang et~al.(2021)Zhang, Hao, and Hu]{19}
W.~Zhang, C.~Hao, and Z.~Hu.
\newblock Retrieval of rainstorm similarity system based on deep learning.
\newblock \emph{Procedia Computer Science}, 183:\penalty0 152--159, 2021.
\newblock \doi{10.1016/j.procs.2021.02.044}.

\bibitem[Fang et~al.(2021)Fang, Xue, Shen, and Sheng]{20}
W.~Fang, Q.~Xue, L.~Shen, and V.S. Sheng.
\newblock Survey on the application of deep learning in extreme weather prediction.
\newblock \emph{Atmosphere}, 12\penalty0 (6):\penalty0 661, 2021.
\newblock \doi{10.3390/atmos12060661}.

\bibitem[Liu et~al.(2022)Liu, Mao, Wu, Feichtenhofer, Darrell, and Xie]{21}
Z.~Liu, H.~Mao, C.Y. Wu, C.~Feichtenhofer, T.~Darrell, and S.~Xie.
\newblock A convnet for the 2020s.
\newblock In \emph{Proceedings of the IEEE/CVF Conference on Computer Vision and Pattern Recognition}, pages 11976--11986, 2022.

\bibitem[Xu et~al.(2021)Xu, Zhang, Zhang, and Tao]{22}
Y.~Xu, Q.~Zhang, J.~Zhang, and D.~Tao.
\newblock Vitae: Vision transformer advanced by exploring intrinsic inductive bias.
\newblock In \emph{Advances in Neural Information Processing Systems}, volume~34, pages 28522--28535, 2021.

\bibitem[Thuemmel et~al.(2023)Thuemmel, Karlbauer, Otte, Zarfl, Martius, Ludwig, Scholten, Friedrich, Wulfmeyer, Goswami, and Butz]{23}
J.~Thuemmel, M.~Karlbauer, S.~Otte, C.~Zarfl, G.~Martius, N.~Ludwig, T.~Scholten, U.~Friedrich, V.~Wulfmeyer, B.~Goswami, and M.V. Butz.
\newblock Inductive biases in deep learning models for weather prediction.
\newblock \emph{arXiv (Cornell University)}, 2023.
\newblock \doi{10.48550/arxiv.2304.04664}.

\bibitem[Ardelean et~al.(2023)Ardelean, Coporîie, Ichim, Dînşoreanu, and Mureșan]{24}
E.~Ardelean, A.~Coporîie, A.~Ichim, M.~Dînşoreanu, and R.~C. Mureșan.
\newblock A study of autoencoders as a feature extraction technique for spike sorting.
\newblock \emph{PLOS ONE}, 18\penalty0 (3):\penalty0 e0282810, 2023.
\newblock \doi{10.1371/journal.pone.0282810}.

\bibitem[Bank et~al.(2023)Bank, Koenigstein, and Giryes]{25}
D.~Bank, N.~Koenigstein, and R.~Giryes.
\newblock Autoencoders.
\newblock In \emph{Machine Learning for Data Science Handbook: Data Mining and Knowledge Discovery Handbook}, pages 353--374. 2023.

\bibitem[Tschannen et~al.(2018)Tschannen, Bachem, and Lucic]{26}
M.~Tschannen, O.~Bachem, and M.~Lucic.
\newblock Recent advances in autoencoder-based representation learning.
\newblock \emph{arXiv preprint arXiv:1812.05069}, 2018.

\bibitem[Van Den~Oord and Vinyals(2017)]{27}
A.~Van Den~Oord and O.~Vinyals.
\newblock Neural discrete representation learning.
\newblock In \emph{Advances in Neural Information Processing Systems}, volume~30, 2017.

\end{thebibliography}
\end{document}